\documentclass{article}

\PassOptionsToPackage{numbers}{natbib}

\usepackage[final]{neurips_2019}
\usepackage[title]{appendix}

\usepackage[utf8]{inputenc} 
\usepackage[T1]{fontenc}    
\usepackage{hyperref}       
\usepackage{url}            
\usepackage{booktabs}       
\usepackage{nicefrac}       
\usepackage{microtype}      

\usepackage{longtable}
\usepackage{adjustbox}
\usepackage[load-configurations=version-1]{siunitx} 

\usepackage{amsmath,amsfonts,amssymb}

\usepackage{makecell}
\usepackage[symbol]{footmisc}

\renewcommand{\thefootnote}{\arabic{footnote}}
\usepackage{xcolor}
\usepackage{soul}
\newcommand{\mathcolorbox}[2]{\colorbox{#1}{$\displaystyle #2$}}

\renewcommand{\footnotesize}{\scriptsize}

\def\mybar#1{
{\color{cyan}\rule{#1mm}{8pt}}}

\title{Assigning Medical Codes at the Encounter Level\\
by Paying Attention to Documents}

\author{Han-Chin Shing \\
       {\small Dept of Computer Science} \\
       {\small Univ of Maryland, College Park } \\
       \texttt{shing@cs.umd.edu}
       \And
       Guoli Wang \\
       {\small 3M Health Information Systems} \\
       Silver Spring, MD, USA \\
       \texttt{gwang11@mmm.com}
       \And
       Philip Resnik \\
       {\small Linguistics/UMIACS} \\
       {\small Univ of Maryland, College Park} \\
       \texttt{resnik@umd.edu}
       } 
\newcommand{\psrcomment}[1]{}

\newcommand{\guolicomment}[1]{}

\newcommand{\hancomment}[1]{}

\newcommand{\suggestignore}[1]{}

\newcommand{\ignore}[1]{}

\begin{document}

\maketitle

\vspace{-1em}

\begin{abstract}
The vast majority of research in computer assisted medical coding focuses on coding at the document level, but a substantial proportion of medical coding in the real world involves coding at the level of clinical encounters, each of which is typically represented by a potentially large set of documents.  We introduce encounter-level document attention networks, which use hierarchical attention to explicitly take the hierarchical structure of encounter documentation into account. Experimental evaluation demonstrates improvements in coding accuracy as well as facilitation of human reviewers in their ability to identify which documents within an encounter play a role in determining the encounter level codes.
\end{abstract}

\section{Introduction}

\textit{Medical coding} translates unstructured information about diagnoses, treatments, procedures, medications and equipment into alphanumeric codes\ignore{, such as International Classification of Diseases (ICD) codes or Current Procedural Terminology (CPT) codes,} for billing\ignore{or insurance} purposes.
\ignore{To correctly interpret this information, experienced professionals (known as medical coders)}Coding is challenging and expensive, requiring high-expertise professionals,\ignore{Experienced medical coders  involved in the process of medical coding. However, this can be expensive due to the large amount of medical text that needs to be processed, a high degree of expertise is required,}
and even experienced coders frequently disagree with each other \citep{resnik2006using}.
Increasingly, computer assisted coding (CAC) is used to help address these issues by automatically suggesting medical codes, generally within a workflow that supports subsequent human review to ensure that codes are correct or to make revisions.

\ignore{
Significant progress has been made on accurate code assignment and on presenting results in a way that coders can interpret (e.g.~\citep{Mullenbach2018ExplainablePO, baumel2018multi}), and auto-suggestion has been shown to improve inter-coder consistency~\citep{resnik2006using}.
}

The vast majority of relevant literature focuses on automatic code assignment at the document level, such as radiology reports \citep[e.g.][]{farkas2008automatic} or discharge summaries \citep[e.g.][]{perotte2013diagnosis}. However, in many settings codes are assigned not to individual documents, but to an entire medical \emph{encounter},\ignore{i.e. an interaction between a patient and healthcare provider,} such as a patient visit to a hospital.
Encounter-level documentation often involves multiple documents~\citep{o2005measuring},
and the relationship between the encounter-level codes and the unstructured information in the documents is indirect --- so the standard approach, treating coding as a well understood kind of text classification problem~\citep[e.g.][]{Pang:2002:TUS:1118693.1118704, Wang:2012:BBS:2390665.2390688,yang2016hierarchical}, does not map naturally to document \textit{collections}.
One obvious solution, using document-level models and then merging their predictions into encounter codes, immediately runs up against a lack of training data: medical coders do not identify which documents are the ``source'' for each encounter code. In addition, merging document-level codes involves non-trivial interactions, e.g. specific codes suppressing more general codes~\citep{o2005measuring}. 

In this paper, we instead focus on training an \textit{encounter-level} model directly.
One straightforward approach would be to aggregate (via sum or average) all document features into a single encounter feature set, but this would be noisy, as the signal of the targeted medical code is diluted when irrelevant documents are also included. It also fails to address the crucial problem of \emph{interpretability}: human coders reviewing auto-suggested codes need to relate proposed encounter codes back to document-level evidence.\footnote{Interpretability is also important from a technical perspective, to identify problems in the prediction model.}
We therefore introduce a new approach to encounter-level coding, observing that its structure is essentially hierarchical, progressing from textual evidence up to documents, and from there to entire encounters\ignore{ where the labels are to be found}. Our Encounter-Level Document Attention Network (ELDAN) applies the key insights of hierarchical attention networks~\citep[HAN,][]{yang2016hierarchical},
enabling the model to
identify which documents are most relevant in encounters as driven by the encounter-level task.  
We obtain positive results for encounter-level labeling in comparison to a strong, realistic baseline, and also show that the resulting weighting helps coders identify which documents are likely sources for a code.

\vspace{-0.5em}

\section{Related Work}

\vspace{-0.5em}

\ignore{
The automatic identification of medical codes dates back at least to~1973, when~\citet{dinwoodie1973automatic} proposed a dictionary matching method based on clinical code descriptions. Since the 1990s, a growing literature has introduced natural language processing techniques to address the task of automatic clinical coding using the unstructured text data~\citep[][ and many others]{resnik2006using,Pestian2007AST,Zhang2017EnhancingAI,Mullenbach2018ExplainablePO}. Many of these studies, however, focus on limited categories of codes, such as variations of pneumonia, or only analyze specific subsets of clinical documents, such as chest radiology and discharge summary~\citep{stanfill2010systematic}. Progress on this problem using state of the art techniques has also been hampered significantly by the broader research community's limited access to large, shareable datasets~\citep{resnikDataRant2018}.
}
\ignore{
More recently, particularly thanks to the availability of a large public dataset, MIMIC III~\citep{johnson2016mimic}, machine learning work on medical coding has moved toward a more comprehensive approach to the problem.  At the same time, 
}
Most work in CAC is limited to discharge summaries~\citep[e.g.][]{perotte2013diagnosis,stanfill2010systematic}, which are assumed to condense information about a patient stay. This can be problematic: \citet{kripalani2007deficits}\ignore{reviewing~73 published studies investigating hospital communication and information transfer}  find high rates of information missing from discharge summaries,\ignore{including median missing-information rates of 10.5\% for physical findings, 38\% for diagnostic test results, and,} notably 17.5\% missing the main diagnosis.\ignore{, which would therefore need to be identified from other documentation in the encounter.}
In addition, for outpatient encounters discharge summaries are rarely a part of the record.\footnote{\ignore{An expert medical coder observes that this makes sense, since for}The patient is generally not admitted to the facility, and thus will not be discharged.}

Deep learning models have been applied to CAC, \ignore{viewing it as part of an assisted coding process and}some exploiting attention mechanisms to support explainability \citep{baumel2018multi,Mullenbach2018ExplainablePO,shi2017towards}. Crucially, however, these all look solely at the discharge summary. To our knowledge, our work is the first to investigate the hierarchical structure of the encounter as a whole.
Our work draws inspiration from \citet{yang2016hierarchical}, who use a hierarchical word-to-sentence-to-document architecture in sentiment analysis.\ignore{They use a two-level attention mechanism that learns to pay attention to specific words in a sentence to form a sentence representation, and at the next higher level to weight specific sentences in a document in forming a document representation.}~Our own multi-level architecture 
progresses from sparse document features to dense document vectors to encounters.\ignore{work moves a level up the representational hierarchy, learning also to weight documents to form encounter representations. Instead of building the representation from the word level, our work}
Domain-informed feature extraction using 
subject matter knowledge and resources, e.g. UMLS,
\ignore{allows us to incorporate expert knowledge, helps alleviate the problem of out-of-vocabulary and uncommon abbreviation terms often found in clinical notes, and}
permits a shallower network, therefore requiring less training data, which is important in a setting where many codes are rare.


\vspace{-0.5em}

\section{Datasets}
\label{sec:cohort}\label{sec:dataset}

\vspace{-0.5em}

\hancomment{This section is for Guoli to write, but here are some info that would be very nice if it can make it in: 

(1) the corpus is outpatient data (as inpatient data does not have CPT code), 

(2) how the feature extraction is being done, like using UMLS or rule-based approach to identify concept etc..., and possibly mentioning that we have 775330 features?
}

We used outpatient procedure (CPT) coding production data internal to 3M Health Information Systems, a leading provider of CAC solutions, sampled from multiple hospital sites.
Our dataset includes 463,866 coded encounters containing 1,390,605 documents, with 31\% of encounters containing a single document; in the remainder, there are an average of 3.91~documents per encounter.
We generated a random 80-10-10 training/tuning/evaluation split by encounter ID. Coding exists only at the encounter level, with no indication of which codes are associated with which document(s).\footnote{To eliminate risk of inappropriate protected health information (PHI) transmission even internally within 3M HIS, once documents were selected, they were immediately converted from their original form to feature vectors containing UMLS CUIs (Concept Unique Identifiers) and 3M HIS
internal numeric concept identifiers as well as words or phrases  (775,330 unique features) for all downstream machine learning development and experimentation. 
No PHI contributed to the features used to represent documents.}

In addition, to assess the value of document-level attention in identifying which documents are responsible for encounter codes (for facilitating human code review) we extracted a separate dataset from production data, comprising 393 encounters.\footnote{To eliminate possible leakage across experiments these do not overlap with the first set.} \psrcomment{Can we verify that none of the \emph{patients} overlap? Ideally in the train/dev/test split above also? This is a much stronger statement than just saying the encounters do not overlap}For each encounter, experienced medical coders annotated \emph{document}-level codes corresponding to the encounter-level coding. Specifically, coders were instructed to read through all the documents contained in the encounter, and assign a code from the encounter level to the document if (and only if) it contains sufficient evidence for assigning the code. Note this means the same code can be assigned to multiple documents within the encounter.

\vspace{-0.5em}

\section{Model: Encounter-Level Document Attention Network}
\label{sec:model}

\vspace{-0.5em}

The overall architecture of Encounter-Level Document Attention Networks (ELDAN)
consists of three parts: (1)~a document-level encoder that turns sparse document features into dense document features using an embedding layer followed by two fully connected layers, (2)~a document-level attention layer that draws inspiration from~\citet{yang2016hierarchical}, and (3)~an encounter-level encoder using a fully connected layer. See Appendix~\ref{ap:model} for full description.

\ignore{As multiple codes are often associated with an encounter,} Encounter-level coding can be considered a multi-label classification problem. \hancomment{Added justification for our decision to train one-vs-all binary}
We decompose the problem into multiple one-vs-all binary classification problems, each targeting one code\ignore{$c_t \in C$, the set of all codes}, which adds flexibility for use cases where codes of interest could vary across sites or even dynamically, and also facilitates comparing code-specific document attention learned from the model to document annotations labeled by medical coders, in our evaluation below.\footnote{We also plan to explore ELDAN with multi-label classification in future work.} 

When a code is assigned to an encounter, it does not imply that all its documents contain evidence for that code. Directly summing or averaging all the encounter's dense document representations \ignore{in that encounter, $\{d_{i,1}, d_{i,2}, \cdots, d_{i,m}\}$, we} will typically capture irrelevant information, diluting the signal for the presence of the code. Instead, ELDAN computes a weighted average, where more relevant documents receive more attention. This is calculated by comparing the dense document representation to a learnable attention vector, after passing through a fully connected-layer and a non-linear layer (see Appendix~\ref{ap:model}, especially eqs. \ref{eq:transformation_ap}-\ref{eq:avg_ap}).

\vspace{-0.5em}

\section{Experiments} 

\vspace{-0.5em}

\ignore{Our first validation experiment tests ELDAN's effectiveness for predicting encounter-level codes. The second looks at the value of document-level attention from ELDAN as a prediction of which documents in the encounter can be considered the ``source'' for the encounter-level codes.}

\ignore{
We first talk about how the training is set up and the implementation innovation \hancomment{might be too strong, as I used many existing pytorch modules.} needed to scale ELDAN to 150 codes, and then talk about the experiment designs.
}

\paragraph{Evaluating Encounter-Level Code Prediction.}
We train two \textsc{ELDAN} models. One is a standard \textsc{ELDAN} model (Section~\ref{sec:model} and Appendix~\ref{sec:model_training}). The other, which we refer to as \textsc{ELDAN+transfer}, includes a simple but effective enhancement for handling rare codes, since, when the code is rare, training a deep one-vs-all network can be challenging. To address this issue, we use a na\"ive transfer learning technique that initializes the embedding layer \ignore{($W_{Embedding}$)}with that of a trained model on a more frequent code.\footnote{See Appendix~\ref{sec:model_training}, under Training Details. We call this na\"ive as it is clearly not optimal nor novel, but the results demonstrate a potentially promising direction for training classifiers for rare medical codes in settings where a single multi-label classifier may be less desirable for other reasons, as discussed above.}
We measure performance in standard fashion using the F1 score.

We regard Yang et al.'s \citep{yang2016hierarchical} non-attention hierarchical network (HN-AVE in their paper) as a strong baseline since,
in experiments across six document classification datasets, they demonstrated that it substantially outperformed a range of typical baselines lacking hierarchy; these included, for example, bag of words, SVM, LSTM, and CNN classifiers. \suggestignore{(They then demonstrated consistent further improvement for hierarchical network modeling with attention.) }Analogously, we define \textsc{ELDN} (encounter level document network) as a baseline that simply averages documents rather than using attention.\footnote{Note that most prior methods for medical coding base the prediction on a single discharge summary (which is rarely present in outpatient encounters), and are thus not applicable as baselines in our setting.} \suggestignore{Specifically, an equally weighted average is used to combine dense document representations, $\{d_{i,1}, d_{i,2}, \cdots, d_{i,m}\}$, into an encounter representation, $e_i$. }




\paragraph{Evaluating Relevant-Document Prediction against Human Judgments.} To evaluate the extent to which document attention learned by \textsc{ELDAN} matches human medical coders' judgments about the documents relevant for coding the encounter, we apply our trained models to our second dataset. Recall that this is a separate set of 393~encounters for which a team of experienced medical coders annotated codes at the document level.
We calculate \textit{document-level F1-score} by treating document attention learned from \textsc{ELDAN} as the prediction of which documents are the ``source'', and comparing this to medical coders' ground truth --- see Appendix~\ref{ap:doc_f1} for an illustration. Note that this is different from the encounter-level F1 scores used to evaluate encounter-level code prediction.

To determine which \emph{documents} are predicted to contain targeted codes (therefore relevant for human code review of the encounter-level coding), we pass the annotated dataset through the \textsc{ELDAN} model trained for encounter-level code prediction, with no further tuning or training. We then use a selection strategy that takes the attention scores of all the documents in an encounter\ignore{(which can be viewed as a multinomial distribution),} and marks all documents that are strictly larger then half the maximum attention score as containing the targeted code. \suggestignore{Note we did not tune this heuristic strategy on the evaluation dataset: the $0.5$ threshold was determined by the observation on the development set that attention scores are often skewed, but sometimes there exist two or more documents with similar high scores.}Since a baseline to compare with document-level attention can be non-trivial to implement, in the spirit of having a chance-adjusted measure, we compare with a baseline obtained by randomly generating attention scores from a uniform distribution on the documents within an encounter, then following the same selection strategy as in ELDAN's document attention selection. The chance baseline is run~500 times to reduce the noise level.

\vspace{-0.5em}

\section{Results and Discussion} 

\vspace{-0.5em}



\paragraph{Results Evaluating Encounter-Level Code Prediction.} 
\textsc{ELDAN} consistently outperforms the baseline\ignore{\textsc{ELDN}, the version of \textsc{ELDAN} without attention,} for 17 of the most frequent 20 codes (Fig.~\ref{tb:top20_svm}, left).
To show the trend across the full range of codes we macro-average every 10 codes from most frequent to least frequent (Fig.~\ref{tb:top20_svm}, right). \textsc{ELDAN} with or without na\"ive transfer learning consistently outperforms \textsc{ELDN}, even for extremely rare codes ($<0.1\%$). As codes become rarer, \textsc{ELDAN+transfer} tends toward outperforming \textsc{ELDAN} more substantially; see increasing trend for $\Delta$\textsc{ELDAN}. This improvement can be explained by viewing the embedding layer as a vector space model that maps sparse features that are extracted from the document (such as medical concepts, UMLS CUIs) to a dense representation, which can be effective for bootstrapping the training of rare codes.

\begin{figure}[ht]
\centering
\caption{\underline{\emph{Left}}: Encounter-level F1-scores of the 20 most frequent CPT codes.\ \#Docs is the average number of documents found in the encounters that contain the code; prevalence is the percentage of all encounters that contain that code. \underline{\emph{Right}}: Macro average of encounter-level F1 scores for every 10 codes (from most to least frequent). $\Delta$\textsc{ELDAN} $=  \textsc{ELDAN+transfer}  - \textsc{ELDAN}$.
\label{tb:top20_svm} \label{tb:average}
}
\vspace{0.5em}
\resizebox{1.0\columnwidth}{!}{%
\begin{tabular}{l|rr|r|rr}
\toprule
{CPT Codes} &  \#Docs & Prevalence & \textsc{ELDN} &    \textsc{ELDAN} & \thead{\textsc{ELDAN}\\\textsc{+transfer}}  \\
\midrule
43239   &       3.13 &     4.15\% &       84.59 &  \bf{86.21} &          84.93 \\
45380   &       2.78 &     3.56\% &       72.68 &  \bf{75.14} &          74.02 \\
45385   &       2.75 &     2.44\% &       71.33 &  \bf{72.33} &          70.31 \\
66984   &       2.51 &     1.90\% &       92.15 &       92.87 &     \bf{93.00} \\
45378   &       2.40 &     1.89\% &       62.67 &       65.45 &     \bf{67.57} \\
12001   &       2.20 &     1.60\% &  \bf{46.96} &       44.74 &          43.62 \\
12011   &       2.35 &     1.19\% &       41.03 &       42.12 &     \bf{43.30} \\
29125   &       2.85 &     1.05\% &       52.32 &  \bf{56.50} &          54.10 \\
10060   &       2.09 &     1.00\% &       45.15 &       48.73 &     \bf{52.25} \\
69436   &       3.01 &     0.96\% &       83.30 &       85.18 &     \bf{88.32} \\
12002   &       2.60 &     0.92\% &       25.53 &       28.36 &     \bf{28.43} \\
59025   &       1.86 &     0.92\% &  \bf{73.82} &       69.00 &          67.73 \\
11042   &       3.20 &     0.88\% &       64.38 &       63.45 &     \bf{66.86} \\
47562   &       4.36 &     0.80\% &       70.74 &       76.25 &     \bf{77.67} \\
62323   &       2.10 &     0.79\% &       61.17 &       57.07 &     \bf{64.25} \\
\midrule
\midrule
Average &       2.62 &            &       58.02 &       60.40 &     \bf{61.26} \\     
\bottomrule
\end{tabular}
\ \ \ \ \ \ 
\begin{tabular}{r|r|r|rr|r}
\toprule
{Average} & Prevalence &   \textsc{ELDN} &       \textsc{ELDAN} & \thead{\textsc{ELDAN}\\\textsc{+transfer}} & $\Delta$\textsc{ELDAN} \\
\midrule
1st to 10th    &     1.97\% &  65.22 &       66.93 &     \bf{67.14} &   0.22 \\
11st to 20th   &     0.78\% &  50.82 &       53.87 &     \bf{55.38} &   1.50 \\
21st to 30th   &     0.51\% &  55.93 &  \bf{63.07} &          62.23 &  -0.85 \\
31st to 40th   &     0.40\% &  44.93 &       51.92 &     \bf{55.24} &   3.32 \\
41st to 50th   &     0.30\% &  32.08 &       38.61 &     \bf{39.35} &   0.74 \\
51st to 60th   &     0.26\% &  33.83 &       38.80 &     \bf{39.10} &   0.30 \\
61st to 70th   &     0.23\% &  28.37 &       35.05 &     \bf{36.62} &   1.56 \\
71st to 80th   &     0.21\% &  25.66 &       30.62 &     \bf{32.93} &   2.31 \\
81st to 90th   &     0.18\% &  34.92 &       42.03 &     \bf{43.26} &   1.23 \\
91st to 100th  &     0.16\% &  24.54 &       29.06 &     \bf{31.32} &   2.25 \\
101st to 110th &     0.14\% &  25.15 &       33.17 &     \bf{34.57} &   1.40 \\
111st to 120th &     0.12\% &  24.87 &       31.74 &     \bf{32.84} &   1.09 \\
121st to 130th &     0.11\% &  18.14 &       24.10 &     \bf{28.09} &   3.99 \\
131st to 140th &     0.10\% &  20.39 &       28.53 &     \bf{32.21} &   3.68 \\
141st to 150th &     0.08\% &  26.93 &       33.13 &     \bf{40.94} &   7.82 \\
\bottomrule
\end{tabular}
}
\vspace{-0.2em}
\end{figure}

\begin{table}[ht]
\centering
\caption{Document-level F1-score calculated by comparing document attention from \textsc{ELDAN} and human coders on 20 CPT codes. \#enc is the number of encounters that contains the code. \#doc is the number of documents within those encounters. \#source is the number of documents being labeled by human coders as the source documents for the code.
Attention (from \textsc{ELDAN}) and Chance both report document-level F1-score, and Diff is the difference between them.}

\resizebox{1.0\columnwidth}{!}{%

\resizebox{0.40\columnwidth}{!}{%
\begin{tabular}{l|rrr|rr|r}
\toprule
{CPT Codes} &  \#enc &  \#doc&  \#source &  Attention &  Chance &  Diff \\
\midrule
43239 &               8 &                       19 &              9 &      88.89 &   59.22 & 29.67 \\
45380 &               5 &                       11 &              5 &      90.91 &   56.47 & 34.44 \\
45385 &               6 &                       13 &              8 &      85.71 &   67.52 & 18.20 \\
66984 &               7 &                       13 &              7 &     100.00 &   68.65 & 31.35 \\
45378 &              10 &                       20 &             11 &      90.91 &   67.44 & 23.47 \\
12001 &               1 &                        3 &              1 &     100.00 &   45.63 & 54.37 \\
12011 &               3 &                        8 &              3 &      57.14 &   54.30 &  2.85 \\
29125 &               2 &                        9 &              4 &      72.73 &   50.91 & 21.81 \\
10060 &               4 &                        9 &              6 &     100.00 &   71.65 & 28.35 \\
69436 &               7 &                       18 &              8 &      87.50 &   60.54 & 26.96 \\
\bottomrule
\end{tabular}
}

\hspace{0.5em}

\resizebox{0.40\columnwidth}{!}{%
\begin{tabular}{l|rrr|rr|r}
\toprule
{CPT Codes} &  \#enc &  \#doc&  \#source &  Attention &  Chance &  Diff \\
\midrule
12002 &               4 &                       13 &              6 &      92.31 &   56.02 & 36.29 \\
59025 &             0 &                      0 &            0 &        - &     - &   - \\
11042 &               5 &                       23 &             16 &      58.06 &   64.89 & -6.82 \\
47562 &               1 &                        5 &              3 &     100.00 &   57.62 & 42.38 \\
62323 &               5 &                       11 &              7 &      87.50 &   69.85 & 17.65 \\
64483 &               3 &                        8 &              4 &     100.00 &   58.07 & 41.93 \\
43235 &               6 &                       18 &              6 &      83.33 &   45.19 & 38.15 \\
20610 &               5 &                        9 &              5 &     100.00 &   72.25 & 27.75 \\
49083 &              10 &                       27 &             13 &      85.71 &   60.21 & 25.50 \\
51702 &               2 &                        2 &              2 &     100.00 &  100.00 &  0.00 \\
\bottomrule
\end{tabular}
}
}

\label{tb:attention}

\vspace{-0.2em}

\end{table}

\paragraph{Results Evaluating Relevant-Document Prediction against Human Judgments.}
Table~\ref{tb:attention} shows document-level F1-score for the most frequent~20 encounter-level codes, with surprisingly strong results: 100\% F1-score on 7 out of 19 available codes.\footnote{Note that as the dataset is smaller and disjoint from the training dataset, codes can be missing (such as code 59025).} However, even chance performance could be good if the number of possible documents to assign credit to is very small.\footnote{As an extreme case, performance for code 51072 is evaluated on two encounters, each of which contains only a single document (Table~\ref{tb:attention}), though this is atypical.} Therefore we compare
to the chance baseline. ELDAN is consistently better, usually by a large margin.\footnote{Except for one code. Improvement is significant at p < .05 using a one-sample t-test comparing the population mean of average F1 over the 500 chance baseline runs against the document-level F1 obtained using the document attention model.} These results support the conclusion that ELDAN's document attention is effective in identifying signal from ``source'' documents for the targeted code --- crucially, without training on document-level annotations.

\ignore{
 number of documents in those encounters that contain the code, as well as the number of ``source'' documents labeled by human coders in Table~\ref{tb:attention}. It is evident that for many high performing codes, the strong results hold.
}
 \ignore{For the one code that it does worse, the difference is relatively low, and could be due to variance resulted from evaluating on a small dataset.} 

\vspace{-0.2em}

\section{Conclusions and Future Work}

\vspace{-0.2em}

In this paper, we have introduced a new approach to encounter-level coding that explicitly takes the hierarchical structure of encounter documentation into account.
Experimental validation of the model shows that it improves coding accuracy against a strong baseline, and also supports the conclusion that its assignment of document-level attention would provide value in helping human coders to identify document-level evidence for encounter-level codes during review. We also found that, in a setting using a set of one-vs-all classifiers, a na\"ive transfer learning approach was surprisingly effective in helping to deal with rare codes.

\ignore{
    \item Implementation-level innovations needed to scale ELDAN up to a real-world number of codes. \hancomment{I am not so sure about this being a strong contribution now, as all other deep learning people just train one multi-label model, instead of 50 binary models, which help with the speed.}
    \item Naive transfer learning \hancomment{(reference to Hal’s work and follow-ons) I read through Hal's work, I think it is probably a stretch to make a connection?}: which is surprising effective for helping with rare codes.
}


In future work, we are particularly interested in exploring the further potential afforded by the assignment to individual documents of credit for encounter-level codes. ELDAN's document-attention can be viewed as a multinomial distribution across documents in an encounter, weighting candidate source documents.  This can be interpreted as a form of weak supervision, and incorporated either via the loss function, or by bringing human judgments back into the loop, e.g. applying active learning to focus on obtaining high quality annotations for valuable subsets of noisily-annotated documents.
An additional focus for future work will be to explore ELDAN with multi-task and multi-label learning, as well as further variants of na\"ive transfer, taking advantage of domain knowledge by grouping codes from the same code family together.

\newpage
\bibliography{main}

\newpage

\begin{appendices}

\section{Dataset Statistics}

\ignore{The experiment dataset includes 463,866 coded encounters containing 1,390,605 documents in the corpus, 31\% (145,518) of the encounters containing a single document. In the remaining 69\% of the dataset, encounter documentation contained an average of 3.91~documents; see }
Table~\ref{tbl:corpus} shows a histogram of encounters that contain a specific number of documents.

\begin{table}[h!]
\centering
\begin{adjustbox}{angle=90,width=0.4\columnwidth}
\begin{tabular}{r|rl}
&  & \# of encounters \\ \midrule
& & \\
\rotatebox[origin=c]{270}{1}  &  145,518 &  \mybar{20.79} \\
\rotatebox[origin=c]{270}{2}  &  113,706 &  \mybar{16.24} \\
\rotatebox[origin=c]{270}{3}  &   72,580 &  \mybar{10.37} \\
\rotatebox[origin=c]{270}{4}  &   45,864 &   \mybar{6.55} \\
\rotatebox[origin=c]{270}{5}  &   28,364 &   \mybar{4.05} \\
\rotatebox[origin=c]{270}{6}  &   17,935 &   \mybar{2.56} \\
\rotatebox[origin=c]{270}{7}  &   12,242 &   \mybar{1.75} \\
\rotatebox[origin=c]{270}{8}  &    8,274 &   \mybar{1.18} \\
\rotatebox[origin=c]{270}{9}  &    5,648 &   \mybar{0.81} \\
\rotatebox[origin=c]{270}{10} &    3,989 &   \mybar{0.57} \\
\rotatebox[origin=c]{270}{11} &    2,863 &   \mybar{0.41} \\
\rotatebox[origin=c]{270}{12} &    2,190 &   \mybar{0.31} \\
\rotatebox[origin=c]{270}{13} &    1,663 &   \mybar{0.24} \\
\rotatebox[origin=c]{270}{14} &    1,217 &   \mybar{0.17} \\
\rotatebox[origin=c]{270}{15} &     997 &   \mybar{0.14} \\
\rotatebox[origin=c]{270}{16} &     816 &   \mybar{0.12} \\
\end{tabular}
\end{adjustbox}

\# of documents in an encounter
\caption{Histogram of the number of documents in an encounter. 
\label{tbl:corpus}}
\vspace{-1em}
\end{table}

\section{Model and Training Details}

\label{ap:model}

\begin{figure}[h!]
  \centering 
  \includegraphics[width=0.7\textwidth]{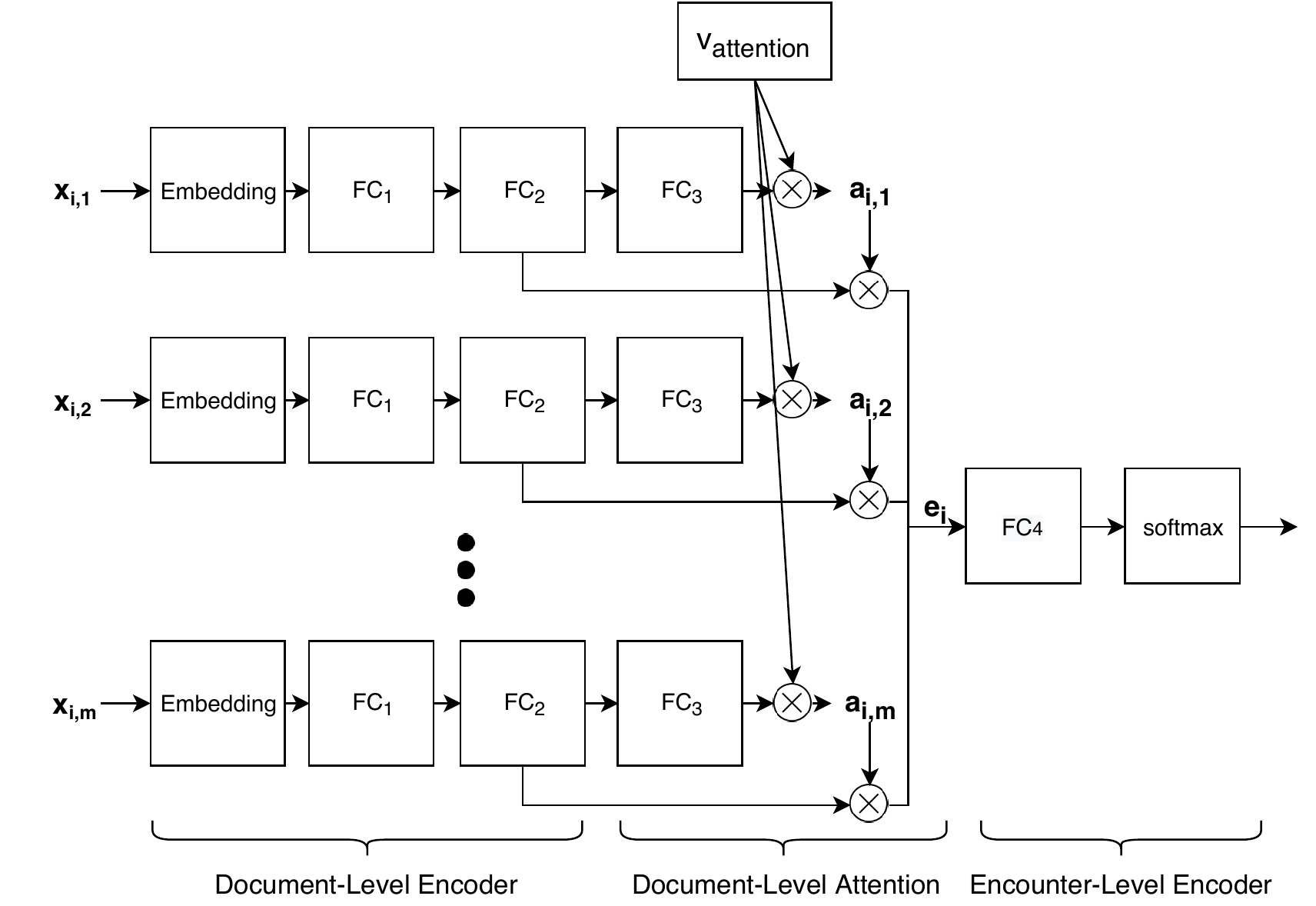} 
  \caption{Architecture of Encounter-Level Document Attention Network (ELDAN)}
  \label{fig:eldan}
\end{figure}

The overall architecture of Encounter-Level Document Attention Networks (ELDAN) is shown in Figure~\ref{fig:eldan}. It 
consists of three parts: (1)~a document-level encoder that turns sparse document features into dense document features, (2)~a document-level attention layer\ignore{ that draws inspiration from~\citet{yang2016hierarchical}}, and (3)~an encounter-level encoder. 

\ignore{As multiple codes are often associated with an encounter, encounter-level coding can be considered a multi-label classification problem. \hancomment{Added justification for our decision to train one-vs-all binary}
For our interest in evaluating the interpretability of ELDAN, we decompose the problem into multiple one-vs-all binary classification problems, with each one targeting one code $c_t \in C = \{c_1, c_2, \cdots, c_K\}$, the set of all codes. This adds flexibility in potential use cases where specific codes of interest could vary across sites or even dynamically, and it makes it easier to compare code-specific document attention learned from the model to document annotations labeled by medical coders.\footnote{We do plan to explore ELDAN with multi-label classification in future work.} }
\psrcomment{Guoli should confirm the ``flexibility in potential use cases'' observation is true and ok to include, or feel free to delete it, keeping just the ``makes it easier to compare...'' piece.}

Let the set of encounters be $E = \{e_1, e_2, \cdots, e_n\}$, and their corresponding labels be $Y = \{y_1, y_2, \cdots, y_n\}$, where $y_i \in \{-1, 1\}$ represents whether the encounter $e_i$ contains the targeted medical code $c_t$. Each encounter $e_i$ comprises multiple documents, and the number of documents that an encounter contains can vary across encounters. Finally, let $x_{i,j}$ and $d_{i,j}$ be the sparse and dense feature vectors that represent document $j$ in encounter $i$, respectively.

\paragraph{Document-Level Encoder.}
\label{sec:d-encode}

The goal of the document-level encoder is to transform a sparse document representation, $x_{i,j}$, into a dense document representation, $d_{i,j}$. The sparse document representation, $x_{i,j}$ is first passed into an embedding layer, to map the 775,330-dimensional sparse document representation into a 300-dimensional vector. It is then followed by two fully-connected layers to produce a dense document representation, $d_{i,j}$.

\vspace{-1em}

\begin{align}
    h_{i,j,0} &= W_{Embedding} x_{i,j} \label{eq:embedding}\\ 
    h_{i,j,1} &= tanh \left ( W_{FC_1} h_{i,j,0} + b_{FC_1} \right ) \\
    d_{i,j} &= tanh \left ( W_{FC_2} h_{i,j,1} + b_{FC_2} \right )
\end{align}

\vspace{-0.5em}

\noindent where $W$ represents the weight matrix, $b$ represents a bias vector, and tanh is the hyperbolic tangent. $h_{i,j,0}$ and $h_{i,j,1}$ are hidden representations of document $j$ in encounter $i$.

\paragraph{Document-Level Attention.}
\label{sec:d-atten}

\ignore{When a medical code is assigned to an encounter, it does not imply that all the documents in the encounter contain evidence for the medical code. If we directly aggregate (whether by summing or averaging) all the dense document representations in that encounter, $\{d_{i,1}, d_{i,2}, \cdots, d_{i,m}\}$, we will typically end up including irrelevant information that dilutes the signal of the presence of medical code. Instead, we might want a weighted average, where the more relevant documents are being paid more attention. }To calculate attention for a document, the dense document representation $d_{i,j}$ is compared to a learnable attention vector, $v_{attention}$, after passing through a fully-connected layer and a non-linear layer. Specifically,

\vspace{-1em}

\begin{align}
        u_{i,j} &= tanh\left ( W_{FC_3} d_{i,j} + b_{FC_3} \right ) \label{eq:transformation_ap}  \\
        a_{i,j} &= \frac{exp\left (u_{i,j}^\top v_{attention}\right )}{\sum_{j=1}^m exp\left (u_{i,j}^\top v_{attention}\right )} \label{eq:atten_ap} \\
        e_{i} &= \sum_{j=1}^m a_{i,j} d_{i,j} \label{eq:avg_ap}
\end{align}

\vspace{-1em}

\noindent where $a_{i,j}$ is the normalized attention score for document $j$ in encounter $i$, and $e_{i}$ is the encounter representation of encounter $i$. As shown in Equation~\ref{eq:atten_ap}, the transformed document representation $u_{i,j}$ is compared with the learnable attention vector $v$ using dot product, and further normalized for the weighted averaging step in Equation~\ref{eq:avg_ap}.

\paragraph{Encounter-Level Encoder.}
\label{sec:e-encode}

Once we have the encounter representation $e_i$, we can predict whether the encounter contains the targeted medical code. Specifically,

\vspace{-1em}

\begin{equation}
    P(\hat{y_i}) = softmax\left ( W_{FC_4} e_{i} + b_{FC_4} \right )
\end{equation}

\vspace{-0.5em}

\noindent Finally, we compare with the ground truth label of encounter $i$ using negative log-likelihood to calculate a loss 
$- log \left ( p(\hat{y_i} = y_i) \right )$ on encounter $i$, where $y_i$ is the ground-truth label. 

\ignore{
\vspace{-1em}
\begin{equation}
\label{eq:loss}
    Loss_i = - log \left ( p(\hat{y_i} = y_i) \right )
\end{equation}
\vspace{-0.7em}
\noindent where $y_i$ is the ground-truth label. 
}

\label{sec:model_training}
\paragraph{Training details.}
Our 80-10-10 dataset split results in 371,092 encounters for training, 46,387 encounters for development/tuning, and 46,387 encounters for testing. Note that no document-level annotations are available. We train models implemented with PyTorch~\citep{paszke2017automatic} on the~150 most frequent codes, using minibatch stochastic gradient descent~\citep{Sutskever:2013:IIM:3042817.3043064} with a minibatch size of 64, learning rate of 0.01, and a momentum of 0.9. Since we are in an imbalanced setting (some medical codes can be extremely rare, see Fig.~\ref{tb:average}), we randomly resampled the training data by assigning different probability to the positive and negative classes so that the ratio of positive encounters and negative encounters is close to $1:6$. No resampling is done for the development set and test set. These hyperparameters were selected based on our results on the development set.

For na\"ive transfer learning, models are trained from the most frequent code to the least frequent. The model for the most frequent code is trained from scratch just like \textsc{ELDAN}. For all the other models, the weight of the $(n)$-th most frequent model's embedding layer ($W_{Embedding}$, see Equation~\ref{eq:embedding}) is first initialized (but not fixed) by that of the $(n-1)$-th most frequent model prior to training.

\section{Document Level F1-score}

\label{ap:doc_f1}

\begin{table}[h!]

\centering
\caption{An illustration of how ELDAN's document attention predictions are evaluated using source documents labeled by human coders. Green (the shading under Human Coders) indicates the ``source'' documents for the encounter-level code (truth), and gray (the shading under Eldan's Document Attention) indicates the documents with high attention (prediction). The bolded documents are the true positives. In this example, the precision is $\frac{tp}{tp + fp}=\frac{3}{3 + 2} = \frac{3}{5}$. The recall is $\frac{tp}{tp + fn}=\frac{3}{3 + 1} = \frac{3}{4}$. The document-level F1 score is thus $\frac{2}{3}$.}
\resizebox{0.50\columnwidth}{!}{%
\begin{tabular}{r|r|r}
\toprule
Encounter & ELDAN's Document Attention & Human Coders\\
\midrule
$enc_1$ & $[\mathcolorbox{lightgray}{\bf{doc_1}}, \mathcolorbox{lightgray}{doc_2}, doc_3]$ & $[\mathcolorbox{green}{\bf{doc_1}}, doc_2, doc_3]$ \\
\midrule
$enc_2$ & $[\mathcolorbox{lightgray}{\bf{doc_4}}]$ & $[\mathcolorbox{green}{\bf{doc_4}}]$ \\
\midrule
$enc_3$ & $[doc_5, \mathcolorbox{lightgray}{\bf{doc_6}}, doc_7, \mathcolorbox{lightgray}{doc_8}]$ & $[\mathcolorbox{green}{doc_5}, \mathcolorbox{green}{\bf{doc_6}}, doc_7, doc_8]$ \\
\bottomrule
\end{tabular}
}

\label{tb:atten-example}
\end{table}

\ignore{In addition to reporting the document-level F1-score, we report, for each code, the number of encounters that contain the code, the number of documents in those encounters, and the number of ``source'' documents labeled by medical coders. Finally,}
To calculate document-level F1-score, we limit our encounters to those that contain the targeted code based on the encounter-level labels, since~(1) there are no annotations for documents that are not in those encounters, and~(2) for ELDAN, attention on the negative encounter can imply negative correlation, which is irrelevant to what we want to evaluate.

\ignore{
\section{Additional Results}

THIS APPENDIX SECTION HAS BEEN RE-INTEGRATED WITH THE MAIN BODY OF THE PAPER

\ignore{
As shown in Table~\ref{tb:top20_svm}, \textsc{ELDAN} consistently outperforms the baseline \ignore{\textsc{ELDN}, the version of \textsc{ELDAN} without attention,} for 17 out of the most frequent 20 codes. This demonstrates the effectiveness of an attention mechanism on the document level. \ignore{This makes sense, as during the medical coding process, some documents can be more important compared to other documents for making a decision on whether to code the specific medical code.}



\begin{table}[h!]
\centering
\caption{Encounter-level F1-scores of 20 most frequent CPT codes. \#Docs is the average number of documents found in the encounters that contain the code. Prevalence is the percentage of all encounters that contain that code.}
\resizebox{0.55\columnwidth}{!}{%
\begin{tabular}{l|rr|r|rr}
\toprule
{CPT Codes} &  \#Docs & Prevalence & \textsc{ELDN} &    \textsc{ELDAN} & \thead{\textsc{ELDAN}\\\textsc{+transfer}}  \\
\midrule
43239   &       3.13 &     4.15\% &       84.59 &  \bf{86.21} &          84.93 \\
45380   &       2.78 &     3.56\% &       72.68 &  \bf{75.14} &          74.02 \\
45385   &       2.75 &     2.44\% &       71.33 &  \bf{72.33} &          70.31 \\
66984   &       2.51 &     1.90\% &       92.15 &       92.87 &     \bf{93.00} \\
45378   &       2.40 &     1.89\% &       62.67 &       65.45 &     \bf{67.57} \\
12001   &       2.20 &     1.60\% &  \bf{46.96} &       44.74 &          43.62 \\
12011   &       2.35 &     1.19\% &       41.03 &       42.12 &     \bf{43.30} \\
29125   &       2.85 &     1.05\% &       52.32 &  \bf{56.50} &          54.10 \\
10060   &       2.09 &     1.00\% &       45.15 &       48.73 &     \bf{52.25} \\
69436   &       3.01 &     0.96\% &       83.30 &       85.18 &     \bf{88.32} \\
12002   &       2.60 &     0.92\% &       25.53 &       28.36 &     \bf{28.43} \\
59025   &       1.86 &     0.92\% &  \bf{73.82} &       69.00 &          67.73 \\
11042   &       3.20 &     0.88\% &       64.38 &       63.45 &     \bf{66.86} \\
47562   &       4.36 &     0.80\% &       70.74 &       76.25 &     \bf{77.67} \\
62323   &       2.10 &     0.79\% &       61.17 &       57.07 &     \bf{64.25} \\
\midrule
\midrule
Average &       2.62 &            &       58.02 &       60.40 &     \bf{61.26} \\     
\bottomrule
\end{tabular}
}

\label{tb:top20_svm}
\vspace{-0.5em}
\end{table}
}
}

\end{appendices}

\end{document}